\documentclass[11pt]{article}

\usepackage[final]{acl}

\usepackage{times}
\usepackage{latexsym}
\usepackage{graphicx}
\usepackage[T1]{fontenc}

\usepackage[utf8]{inputenc}

\usepackage{microtype}

\usepackage{inconsolata}

\usepackage{graphicx}

\title{From NLG Evaluation to Modern Student Assessment in the Era of ChatGPT: The Great Misalignment Problem and Pedagogical Multi-Factor Assessment (P-MFA)}

\author{Mika Hämäläinen and Kimmo Leiviskä \\
  Metropolia University of Applied Sciences \\
  Helsinki, Finland \\
  \texttt{first.last@metropolia.fi} \\}

\begin{document}
\maketitle
\begin{abstract}
This paper explores the growing epistemic parallel between NLG evaluation and grading of students in a Finnish University. We argue that both domains are experiencing a Great Misalignment Problem. As students increasingly use tools like ChatGPT to produce sophisticated outputs, traditional assessment methods that focus on final products rather than learning processes have lost their validity. To address this, we introduce the Pedagogical Multi-Factor Assessment (P-MFA) model, a process-based, multi-evidence framework inspired by the logic of multi-factor authentication. 
\end{abstract}

\section{Introduction}

In recent years, both the field of computational creativity and university pedagogy have faced a growing crisis of evaluation. In creative natural language generation research, \citet{hamalainen-alnajjar-2021-great} articulated the Great Misalignment Problem, which is the disconnection between a system’s problem definition, its implemented method and the evaluation criteria used to assess its performance. This misalignment leads to superficial or misleading conclusions about a system’s success, as the evaluation often fails to measure what the system was designed to achieve.

A similar issue now permeates higher education: the act of grading has become increasingly detached from the authentic learning processes it is meant to assess. As generative models such as ChatGPT\footnote{https://chatgpt.com} can produce fluent, original-looking outputs on demand, educators can no longer be sure whether a student’s submitted work reflects genuine understanding or merely the clever use of an external artificially cognitive tool.

Just as computational creativity systems generate artefacts whose internal reasoning is opaque, students in the AI-saturated learning environment can now present creative products without revealing the intellectual pathway that led to them. In both cases, the evaluator encounters an artefact, such as a poem, a program, an essay or a design, without direct access to the underlying process that produced it. The traditional product-based evaluation paradigm, which assumes a transparent correspondence between output and competence, thus collapses. The opacity of generative systems mirrors the opacity of modern student work: both appear impressive on the surface, yet the link between creation and creator, between performance and understanding, is obscured.

This parallel suggests that the Great Misalignment Problem has re-emerged in pedagogy under new conditions. In education, the “problem definition” corresponds to the intended learning outcomes; the “method” is the student’s learning process; and the “evaluation” is grading. When these three components fall out of alignment - when grades reflect polished submissions rather than genuine cognitive engagement - the integrity of learning assessment is compromised. The university thus faces a dilemma akin to that of computational creativity research: it risks optimizing for the appearance of creativity and competence rather than for the authentic development of these qualities.

To resolve this, both AI research and pedagogy must shift their evaluative focus from product to process. In computational creativity, meaningful evaluation requires attention to how a system produces its output, its generative mechanisms, constraints, and reasoning. Likewise, effective pedagogy must emphasize the learning process itself: reflection, iteration, collaboration, and the student’s evolving relationship with knowledge. In both domains, understanding the process behind the product restores transparency, accountability, and interpretability. The challenge, then, is not merely to detect misuse of generative tools but to redesign evaluation frameworks that value the act of creation—the unfolding of thought—as the true site of learning and creativity.

\section{Related Work}

The emergence of LLMs has received a mixed response among educators; several feel that AI is a threat to students' learning \cite{yu2023reflection,lin2023concerns,ogugua2023academic}, while others focus on it's potential in future of education \cite{lo2023impact,silva2024innovating,morgan2024implementing,macias2024empowering}.

Evaluation of NLG systems, creative and regular, has received a fair share of attention in the past \cite{clark-etal-2021-thats,freitag-etal-2021-experts,howcroft-etal-2020-twenty}. Some researches even highlight that automated evaluation methods such as BLEU are simply not sufficient \cite{reiter2018structured}. Evaluation mainly relies on evaluating the output of such systems rather than taking the creative process into account \cite{hamalainen-alnajjar-2021-human}.

In terms of pedagogy, how people learn and how they should be taught is a relatively well-understood phenomenon. Frameworks such as Bloom's (\citeyear{bloom1956taxonomy}) taxonomy, deep learning (see \citealt{mcgregor2020emerging}) and constructive alignment \cite{biggs1996enhancing} have been widely used. However, there's a big gap between theory and practice, and ultimately, student assessment relies on grading some sort of a final product of learning such as an essay or thesis.

\section{The Great Misalignment Problem in NLG Evaluation}

In their work, \citet{hamalainen-alnajjar-2021-great} identify what they term \textit{the Great Misalignment Problem} in the context of human evaluation in natural language generation (NLG). The core claim is that in much of NLG research that relies on human judgments, there is a systematic misalignment among three key components: (1) how the research problem is defined, (2) how the proposed method or model is formulated and (3) how the human evaluation is conducted. When these three are not tightly aligned, the authors argue, the validity, interpretability and reproducibility of human evaluation outcomes are severely compromised.

The authors support their claim by surveying ten randomly selected papers from ACL 2020 that include human evaluation. In their analysis, they examine (a) whether the problem definition is clearly and narrowly stated, (b) whether the proposed method follows directly from that problem definition, and (c) whether the human evaluation aligns both with the definition and with what the method is intended to model. They report that in only a single case among the ten did all three align. In many cases, the evaluation either ignores aspects of what the method is modeling or tests orthogonal criteria not grounded in the stated problem.

The implications of the Great Misalignment Problem are profound. Because human evaluation may end up measuring something other than what the model is intended to do, the reported improvements or differences in scores cannot reliably be attributed to the proposed method. Instead, they might arise from unintended artifacts, evaluation design biases, evaluator variance or other uncontrolled factors. This undermines claims of advancement, makes comparison across systems less meaningful, and complicates reproducibility. Moreover, the authors point out that human evaluation in NLP is often conducted with insufficient methodological rigor (e.g. vague questions, low numbers of judges and opaque protocols), further exacerbating the misalignment.

To move forward, the authors recommend that NLP researchers take the problem definition seriously and design methods and evaluations so as to maintain alignment. Concretely, they urge narrowing broad, vague problem statements into more precise, measurable sub-tasks; ensuring that modeling decisions correspond to those sub-dimensions; and crafting evaluation questions that directly probe the modeled behavior. They also call for full transparency in evaluation setup (e.g. prompt wording, judge selection and instructions) and suggest that human evaluation practices in NLP could benefit from importation of best practices from fields accustomed to subjective measurement (e.g. social sciences). They do not advocate abandoning human evaluation altogether, but rather reforming it so that it becomes a more trustworthy and interpretable component of NLP research.

\section{LLMs and Teachers' Nightmare}

Not unlike the findings described by \citet{hamalainen-2024-legal}, we have encountered negative teacher narratives in Metropolia University of Applied Sciences regarding LLM tools. There is a lot of fear among teachers that students would use the new technology to cheat and ultimately pass their courses without learning much. On the other hand, there are also teachers who embrace the new technology and actively use it in their teaching.

It is fair to say that LLMs have introduced a radical disruption to the long-standing epistemic contract between teachers and students. Pedagogically, this contract rests on an implicit trust: that the student’s submitted work represents their own intellectual labor and engagement with the learning process. The teacher, in turn, evaluates this work as evidence of learning, understanding and skill development. However, with the advent of LLMs capable of producing contextually appropriate, grammatically flawless and even stylistically distinct texts, this foundational trust is breaking down. Teachers now face the uneasy possibility that a student’s polished essay or thoughtful reflection may be more a testament to prompt engineering than to actual comprehension.

From a pedagogical standpoint, this collapse of evaluative certainty threatens the very rationale of assessment (see \citealt{csahin2024opportunities,fagbohun2024beyond}). If an assignment can be completed without genuine learning, then grades cease to measure educational achievement. They become measures of access to tools and skill in their manipulation. The anxiety many teachers experience arises not merely from the fear of academic dishonesty but from the erosion of pedagogical meaning itself. When outputs can no longer be reliably linked to the cognitive processes they are meant to demonstrate, the educational system loses its anchor: learning becomes performative rather than transformative.

The teacher’s nightmare is not that students are cheating, but that the act of evaluation has become epistemically hollow. Traditional assessment methods such as essays, reports and even project work are predicated on a production model of learning where outputs reflect mental effort. In the LLM era, this assumption no longer holds. The teacher cannot see how the student arrived at a conclusion, whether the reasoning was genuine or algorithmically scaffolded. Consequently, feedback loses precision, as it targets the product rather than the learner’s cognitive or creative process. Pedagogically, this creates a feedback loop of disengagement: students learn to outsource tasks, and teachers, sensing the futility of assessment, may lower expectations or turn toward increasingly mechanistic forms of surveillance.

In essence, LLMs expose the brittleness of output-oriented pedagogy. The crisis they introduce is not technological but epistemological: education must now grapple with redefining what it means to know, learn, and create in a world where human and machine outputs are indistinguishable. Teachers’ frustration, then, is not only emotional but structural. It stems from a system designed for a pre-AI understanding of authorship and agency. The nightmare can only end when pedagogical practices evolve to prioritize the evaluation of the learning process itself over the evaluation of a final product, reclaiming assessment as a shared inquiry into learning rather than a judgment of finished artefacts.

The Great Misalignment Problem offers a valuable lens for rethinking student assessment. By highlighting the dangers of disconnect between problem definition, method and evaluation, the same critique encourages educators to scrutinize whether grading practices truly measure the intended learning outcomes. \textbf{In pedagogy, this means aligning what we want students to learn (problem definition), how they engage in that learning (method), and how we evaluate their understanding (evaluation}).

Rather than focusing solely on the final product, educators can design assessments that make the learning process visible through reflective writing, process logs, peer discussions or iterative project development. In doing so, the evaluation becomes an inquiry into alignment itself: does the student’s process reflect the intended learning goals, and does the teacher’s evaluation capture that process accurately? This alignment-centered approach transforms grading from an act of judgment into an act of dialogue, ensuring that assessment remains meaningful even in an era when the boundary between human and machine creativity is increasingly blurred.



\section{Future of Grading: a P-MFA Approach}

\textbf{We propose a novel framework named Pedagogical Multi-Factor Assessment (P-MFA)} for grading and learning evaluation designed for the age of generative AI. It builds on the logic of multi-factor authentication (see \citealt{ometov2018multi}): just as digital security no longer relies on a single password, educational assessment should not depend on a single artefact such as an exam or essay. P-MFA therefore verifies learning through multiple complementary “factors,” each representing a distinct dimension of competence—what the student knows (knowledge), produces (outputs), can do (application), sustains over time (process continuity), reflects upon (self-evaluation), and connects to real contexts (situated understanding).

By combining these factors, teachers and students co-construct a trustworthy, multi-channel record of learning that is transparent, individualized, and resistant to the misuse of AI. Rather than focusing on control or detection, P-MFA shifts assessment toward alignment: ensuring that what is defined as learning, practiced as learning, and evaluated as learning all converge in an authentic and human-centered educational process.

Theoretically, P-MFA can be understood as a synthesis of constructive alignment and the Great Misalignment Problem. Constructive alignment posits that meaningful learning occurs when intended learning outcomes, teaching activities, and assessment tasks are coherently designed to support one another. The Great Misalignment Problem, in contrast, diagnoses the breakdown of such coherence in research evaluation: when problem definition, method, and evaluation diverge, the resulting claims lose validity. 

P-MFA translates this alignment imperative into pedagogy for the generative-AI era, explicitly designing assessment systems that keep the “problem definition” (learning outcomes), the “method” (student learning processes), and the “evaluation” (grading practices) in continuous dialogue. Where constructive alignment emphasizes curriculum design, P-MFA operationalizes alignment through evidence diversity: multiple, process-anchored factors that ensure the assessment remains faithful to both the intention and the practice of learning. In doing so, P-MFA not only safeguards educational integrity against AI-generated artefacts but also reframes assessment as an interpretive act of maintaining epistemic alignment between what learning is meant to achieve, how it unfolds, and how it is ultimately recognized.

In essence, the P-MFA approach operationalizes the Great Misalignment Problem’s philosophical insight within the classroom. It demands that educators explicitly design for a\textbf{lignment across definition, method, and evaluation}, ensuring that the assessment of learning remains meaningful, transparent, and resilient to technological disruption. By requiring multiple, process-anchored proofs of understanding, P-MFA not only protects the integrity of grading but also redefines it as a dynamic act of alignment, in which learning and evaluation evolve together.

Problem definition in P-MFA is reframed as the articulation of learning outcomes that go beyond static knowledge. Education’s aim is no longer to verify that students can produce isolated outputs but that they can understand, apply, reflect and contextualize their knowledge

The Method of P-MFA corresponds to the pedagogical and learning practices through which these factors are activated. Instead of viewing learning as a linear input–output pipeline, P-MFA promotes iterative, reflective and contextual engagement. Students are not passive respondents to tasks but co-designers of their assessment trajectory, selecting factors that align with their goals and contexts

Finally, Evaluation in P-MFA is no longer an isolated measurement but a process of triangulation. Each factor functions as an evaluative lens that confirms or challenges the authenticity of others. The resulting alignment between what was meant to be learned, how learning occurred, and how it is assessed embodies the very correction the Great Misalignment Problem paper sought in NLP research. When teachers adopt a P-MFA framework, grading transforms from a verdict into an inquiry: a structured investigation into whether the student’s demonstrated process and outputs align with the course’s learning definition. This ensures that evaluation measures authentic engagement rather than algorithmic fluency. Moreover, because AI cannot convincingly reproduce personal reflection or contextual relevance, P-MFA restores pedagogical trust by embedding evaluation in dimensions that remain uniquely human.

\section{Conclusions}

The challenges faced in both computational creativity and contemporary pedagogy converge on a single epistemic issue: the difficulty of evaluating outputs without understanding the processes that produced them. The Great Misalignment Problem revealed how research can lose validity when problem definition, method, and evaluation drift apart, an insight that now illuminates the crisis of grading in the era of generative AI. Our Pedagogical Multi-Factor Assessment (P-MFA) model offers a concrete response to this challenge by embedding assessment within the learning process itself. Through its multi-factor design—combining evidence of knowledge, production, application, continuity, reflection, and context—P-MFA restores alignment between what learning is intended to achieve, how it unfolds, and how it is recognized. In doing so, it reclaims evaluation as a transparent and dialogic practice, reaffirming the role of assessment not as an act of surveillance or verification, but as an interpretive inquiry into human understanding and growth in an age increasingly mediated by machines.


\bibliography{custom}

@inproceedings{hamalainen-alnajjar-2021-great,
    title = "The Great Misalignment Problem in Human Evaluation of {NLP} Methods",
    author = {H{\"a}m{\"a}l{\"a}inen, Mika  and
      Alnajjar, Khalid},
    editor = "Belz, Anya  and
      Agarwal, Shubham  and
      Graham, Yvette  and
      Reiter, Ehud  and
      Shimorina, Anastasia",
    booktitle = "Proceedings of the Workshop on Human Evaluation of NLP Systems (HumEval)",
    month = apr,
    year = "2021",
    address = "Online",
    publisher = "Association for Computational Linguistics",
    url = "https://aclanthology.org/2021.humeval-1.8/",
    pages = "69--74"
}

@article{biggs1996enhancing,
  title={Enhancing teaching through constructive alignment},
  author={Biggs, John},
  journal={Higher education},
  volume={32},
  number={3},
  pages={347--364},
  year={1996},
  publisher={Springer}
}

@article{ometov2018multi,
  title={Multi-factor authentication: A survey},
  author={Ometov, Aleksandr and Bezzateev, Sergey and M{\"a}kitalo, Niko and Andreev, Sergey and Mikkonen, Tommi and Koucheryavy, Yevgeni},
  journal={Cryptography},
  volume={2},
  number={1},
  pages={1},
  year={2018},
  publisher={MDPI}
}

@article{fagbohun2024beyond,
  title={Beyond traditional assessment: Exploring the impact of large language models on grading practices},
  author={Fagbohun, Oluwole and Iduwe, Nwaamaka Pearl and Abdullahi, Mustapha and Ifaturoti, Adeseye and Nwanna, OM},
  journal={Journal of Artificial Intelligence, Machine Learning and Data Science},
  volume={2},
  number={1},
  pages={1--8},
  year={2024}
}

@article{csahin2024opportunities,
  title={Opportunities and Challenges of AI in Educational Assessment},
  author={{\c{S}}ahin, Alper and Thompson, Nathan and Ercikan, Kadriye},
  journal={Journal of Measurement and Evaluation in Education and Psychology},
  volume={15},
  number={Special Issue},
  pages={260--262},
  year={2024},
  publisher={Parantez E{\u{g}}itim Yay{\i}nc{\i}l{\i}k}
}

@inproceedings{hamalainen-2024-legal,
    title = "Legal and Ethical Considerations that Hinder the Use of {LLM}s in a {F}innish Institution of Higher Education",
    author = {H{\"a}m{\"a}l{\"a}inen, Mika},
    editor = "Siegert, Ingo  and
      Choukri, Khalid",
    booktitle = "Proceedings of the Workshop on Legal and Ethical Issues in Human Language Technologies @ LREC-COLING 2024",
    month = may,
    year = "2024",
    address = "Torino, Italia",
    publisher = "ELRA and ICCL",
    url = "https://aclanthology.org/2024.legal-1.5/",
    pages = "24--27"
}

@article{reiter2018structured,
  title={A structured review of the validity of BLEU},
  author={Reiter, Ehud},
  journal={Computational Linguistics},
  volume={44},
  number={3},
  pages={393--401},
  year={2018},
  publisher={MIT Press One Rogers Street, Cambridge, MA 02142-1209, USA journals-info~…}
}

@book{bloom1956taxonomy,
  title={Taxonomy of educational objectives: The classification of educational goals. Handbook 1: Cognitive domain},
  author={Bloom, Benjamin S},
  year={1956},
  publisher={Longman New York}
}

@article{mcgregor2020emerging,
  title={Emerging from the deep: Complexity, emergent pedagogy and deep learning},
  author={McGregor, Sue LT},
  journal={Northeast Journal of Complex Systems (NEJCS)},
  volume={2},
  number={1},
  pages={2},
  year={2020}
}

@inproceedings{hamalainen-alnajjar-2021-human,
    title = "Human Evaluation of Creative {NLG} Systems: An Interdisciplinary Survey on Recent Papers",
    author = {H{\"a}m{\"a}l{\"a}inen, Mika  and
      Alnajjar, Khalid},
    editor = "Bosselut, Antoine  and
      Durmus, Esin  and
      Gangal, Varun Prashant  and
      Gehrmann, Sebastian  and
      Jernite, Yacine  and
      Perez-Beltrachini, Laura  and
      Shaikh, Samira  and
      Xu, Wei",
    booktitle = "Proceedings of the First Workshop on Natural Language Generation, Evaluation, and Metrics (GEM)",
    month = aug,
    year = "2021",
    address = "Online",
    publisher = "Association for Computational Linguistics",
    url = "https://aclanthology.org/2021.gem-1.9/",
    doi = "10.18653/v1/2021.gem-1.9",
    pages = "84--95"
}

@inproceedings{howcroft-etal-2020-twenty,
    title = "Twenty Years of Confusion in Human Evaluation: {NLG} Needs Evaluation Sheets and Standardised Definitions",
    author = "Howcroft, David M.  and
      Belz, Anya  and
      Clinciu, Miruna-Adriana  and
      Gkatzia, Dimitra  and
      Hasan, Sadid A.  and
      Mahamood, Saad  and
      Mille, Simon  and
      van Miltenburg, Emiel  and
      Santhanam, Sashank  and
      Rieser, Verena",
    editor = "Davis, Brian  and
      Graham, Yvette  and
      Kelleher, John  and
      Sripada, Yaji",
    booktitle = "Proceedings of the 13th International Conference on Natural Language Generation",
    month = dec,
    year = "2020",
    address = "Dublin, Ireland",
    publisher = "Association for Computational Linguistics",
    url = "https://aclanthology.org/2020.inlg-1.23/",
    doi = "10.18653/v1/2020.inlg-1.23",
    pages = "169--182"
}

@article{freitag-etal-2021-experts,
    title = "Experts, Errors, and Context: A Large-Scale Study of Human Evaluation for Machine Translation",
    author = "Freitag, Markus  and
      Foster, George  and
      Grangier, David  and
      Ratnakar, Viresh  and
      Tan, Qijun  and
      Macherey, Wolfgang",
    editor = "Roark, Brian  and
      Nenkova, Ani",
    journal = "Transactions of the Association for Computational Linguistics",
    volume = "9",
    year = "2021",
    address = "Cambridge, MA",
    publisher = "MIT Press",
    url = "https://aclanthology.org/2021.tacl-1.87/",
    doi = "10.1162/tacl_a_00437",
    pages = "1460--1474"
}

@inproceedings{clark-etal-2021-thats,
    title = "All That{'}s `Human' Is Not Gold: Evaluating Human Evaluation of Generated Text",
    author = "Clark, Elizabeth  and
      August, Tal  and
      Serrano, Sofia  and
      Haduong, Nikita  and
      Gururangan, Suchin  and
      Smith, Noah A.",
    editor = "Zong, Chengqing  and
      Xia, Fei  and
      Li, Wenjie  and
      Navigli, Roberto",
    booktitle = "Proceedings of the 59th Annual Meeting of the Association for Computational Linguistics and the 11th International Joint Conference on Natural Language Processing (Volume 1: Long Papers)",
    month = aug,
    year = "2021",
    address = "Online",
    publisher = "Association for Computational Linguistics",
    url = "https://aclanthology.org/2021.acl-long.565/",
    doi = "10.18653/v1/2021.acl-long.565",
    pages = "7282--7296"
}

@inproceedings{macias2024empowering,
  title={Empowering Teachers with Usability-Oriented LLM-Based Tools for Digital Pedagogy},
  author={Macias, Melany Vanessa and Kharlashkin, Lev and Huovinen, Leo Einari and H{\"a}m{\"a}l{\"a}inen, Mika},
  booktitle={Proceedings of the 4th International Conference on Natural Language Processing for Digital Humanities},
  pages={549--557},
  year={2024}
}

@article{morgan2024implementing,
  title={Implementing ChatGPT to Support Teachers and Promote Learning},
  author={Morgan, Hani},
  journal={The Clearing House: A Journal of Educational Strategies, Issues and Ideas},
  volume={97},
  number={5},
  pages={125--133},
  year={2024},
  publisher={Taylor \& Francis}
}

@inproceedings{lin2023concerns,
  title={Concerns about using ChatGPT in education},
  author={Lin, Shu-Min and Chung, Hsin-Hsuan and Chung, Fu-Ling and Lan, Yu-Ju},
  booktitle={International conference on innovative technologies and learning},
  pages={37--49},
  year={2023},
  organization={Springer}
}

@article{ogugua2023academic,
  title={Academic integrity in a digital era: Should the use of ChatGPT be banned in schools?},
  author={Ogugua, Divine and Yoon, Seong No and Lee, DonHee},
  journal={Global Business \& Finance Review (GBFR)},
  volume={28},
  number={7},
  pages={1--10},
  year={2023},
  publisher={Seoul: People \& Global Business Association (P\&GBA)}
}

@article{yu2023reflection,
  title={Reflection on whether Chat GPT should be banned by academia from the perspective of education and teaching},
  author={Yu, Hao},
  journal={Frontiers in Psychology},
  volume={14},
  pages={1181712},
  year={2023},
  publisher={Frontiers Media SA}
}

@inproceedings{silva2024innovating,
  title={Innovating for the future: Ai and hrm capabilities for sustainability in higher education},
  author={Cleland Silva, Tricia and H{\"a}m{\"a}l{\"a}inen, Mika},
  booktitle={Academy of Management Annual Meeting},
  volume={2024},
  number={1},
  year={2024}
}

@article{lo2023impact,
  title={What is the impact of ChatGPT on education? A rapid review of the literature},
  author={Lo, Chung Kwan},
  journal={Education sciences},
  volume={13},
  number={4},
  pages={410},
  year={2023},
  publisher={MDPI}
}

\end{document}